\documentclass{article}
\usepackage{spconf,amsmath,epsfig}
\usepackage{amsmath,graphicx}
\usepackage{cite}
\usepackage{amsmath,graphicx}
\usepackage{epstopdf}
\usepackage{xcolor}
\usepackage{amssymb}
\usepackage{pifont}
\usepackage{array}
\usepackage{booktabs}
\usepackage{bm}
\usepackage{multirow}
\usepackage{graphicx}
\usepackage{color}
\usepackage{float}
\usepackage{stfloats}
\usepackage[misc]{ifsym}
\usepackage{stmaryrd}
\usepackage{grffile}
\usepackage{subfig}
\usepackage{makecell}
\usepackage{threeparttable}
\usepackage{enumitem}
\usepackage{enumerate}

\title{ShipSRDet: An End-to-End Remote Sensing Ship Detector Using Super-Resolved Feature Representation}

\name{Shitian~He, Huanxin~Zou*, Yingqian~Wang, Runlin~Li, Fei~Cheng
      \thanks{*Corresponding author: Huanxin Zou}
     }
\address{College of Electronic Science and Technology,\\
         National University of Defense Technology, Changsha 410073, China}

\begin{document}
\maketitle

\begin{abstract}
High-resolution remote sensing images can provide abundant appearance information for ship detection. Although several existing methods use image super-resolution (SR) approaches to improve the detection performance, they consider image SR and ship detection as two separate processes and overlook the internal coherence between these two correlated tasks. In this paper, we explore the potential benefits introduced by image SR to ship detection, and propose an end-to-end network named \textit{ShipSRDet}. In our method, we not only feed the super-resolved images to the detector but also integrate the intermediate features of the SR network with those of the detection network. In this way, the informative feature representation extracted by the SR network can be fully used for ship detection. Experimental results on the HRSC dataset validate the effectiveness of our method. Our \textit{ShipSRDet} can recover the missing details from the input image and achieves promising ship detection performance.

\end{abstract}


\section{Introduction}\label{sec:intro}
Remote sensing (RS) ship detection has attracted extensive attention in recent years due to its large potential in both civilian and military applications (e.g., port management, target surveillance). As a key factor of ship detection, high-resolution (HR) images can provide abundant appearance information and thus introduce improvement to the detection accuracy \cite{shermeyer2019effects}. However, obtaining an HR image posts a high requirement on the satellite sensors and generally results in an expensive cost. Consequently, using image super-resolution (SR) techniques to recover the missing details in RS images has become a popular research topic and has been widely investigated in recent years.

In the area of RS object detection, several methods have been proposed to use image SR as a pre-processing approach to improve the detection accuracy. Dong et al. \cite{dong2020remote} proposed a second-order multi-scale SR network, and demonstrated its effectiveness to RS object detection. Rabbi et al. \cite{rabbi2020small} proposed an edge-enhanced generative adversarial network (EESRGAN) to improve the quality of RS images, and combined EESRGAN and SSD detector in an overall framework to perform end-to-end fine-tuning. Courtrai et al. \cite{courtrai2020small} tailored a GAN-based SR network with a detection network to develop an object-focused detection framework. Note that, although these methods have shown their effectiveness, the benefits introduced by image SR has not been fully exploited since only super-resolved images are fed to the detectors while the informative feature representation extracted by SR networks has been overlooked.

\begin{figure}
\centering
\includegraphics[width=8.2cm]{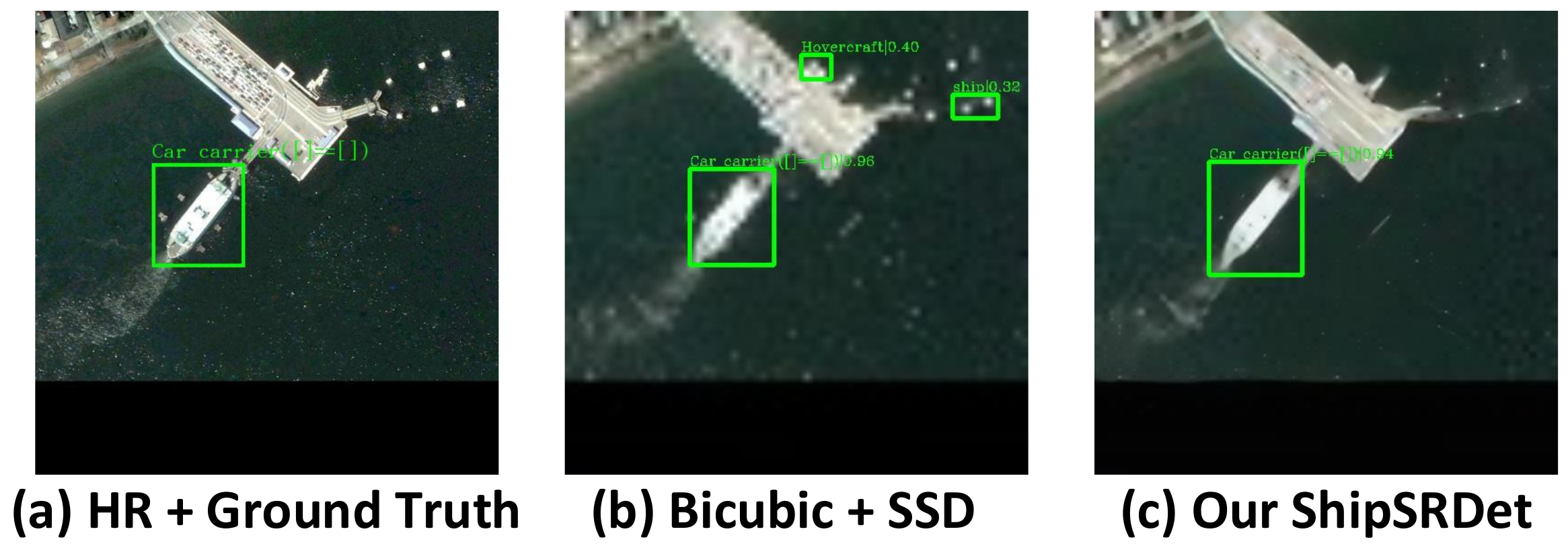}
\vspace{-0.2cm}
\caption{{Visual results achieved by our \textit{ShipSRDet} on the HRSC dataset \cite{HRSC}. Our method recovers missing details in the input image and achieves promising detection performance.}
\label{fig:thumbnail}}
\vspace{-0.2cm}
\end{figure}

Aiming at the aforementioned issue, in this paper, we propose an end-to-end network named \textit{ShipSRDet} to fully use the super-resolved feature representation for RS ship detection. Different from existing SR-based detection methods, we not only feed the super-resolved images to the detector, but also integrate the intermediate features of the SR network with those of the detection network. In this way, more informative cues can be transferred from the SR network to the detection network to enhance the detection performance. Experiments on the HRSC dataset \cite{HRSC} demonstrate the effectiveness of our method. As shown in Fig.~\ref{fig:thumbnail}, our \textit{ShipSRDet} achieves notable improvements on detection performance, and recovers the missing details in the input image.

\begin{figure*}
\centering
\includegraphics[width=17.5cm]{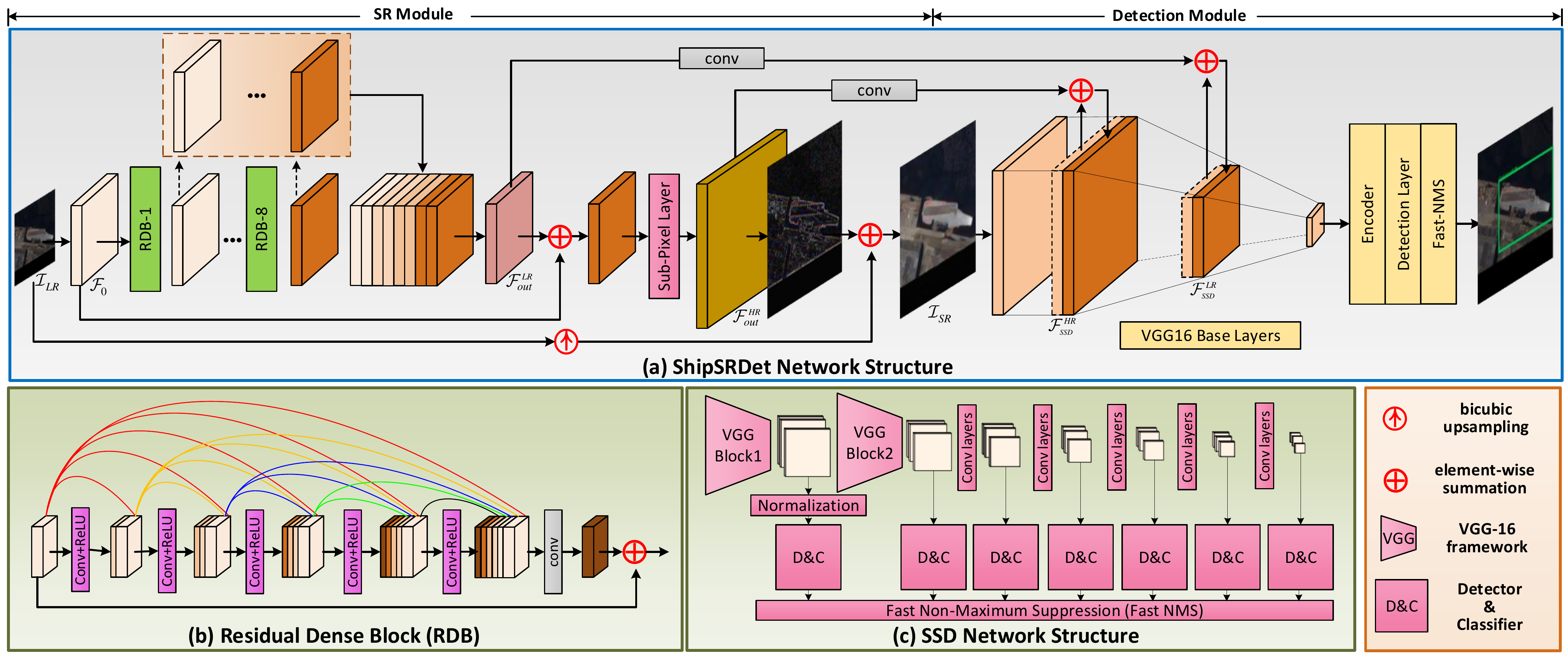}
\vspace{-0.1cm}
\caption{{An overview of our \textit{ShipSRDet}.}
\label{fig:ShipSRDet}}
\end{figure*}

In summary, the contributions of this paper are as follows: \textit{1)} We quantitatively investigate the influence of image quality to RS ship detection. \textit{2)} We proposed a \textit{ShipSRDet} to exploit the potential benefits introduced by HR images and their feature representation to RS ship detection. \textit{3)} Our \textit{ShipSRDet} can recover the missing details in the input images while achieving remarkable detection performance.

\section{Network Architecture}
In this section, we introduce our \textit{ShipSRDet} in details. As shown in Fig.~\ref{fig:ShipSRDet}(a), our \textit{ShipSRDet} consists of two parts including an SR module and a detection module, which will be described in the following subsections, respectively.

\subsection{SR module}
As shown in  Fig.~\ref{fig:ShipSRDet}(a), our SR module takes a medium-low resolution image $\mathcal{I}_{LR}\in\mathbb{R}^{H\times W\times 3}$ as its input to produce an SR image $\mathcal{I}_{SR}\in\mathbb{R}^{\alpha H\times \alpha W\times 3}$ and two intermediate features $\mathcal{F}^{LR}_{out}\in\mathbb{R}^{H\times W\times 64}$ and $\mathcal{F}^{HR}_{out}\in \mathbb{R}^{\alpha H\times \alpha W\times 64}$, where $H$ and $W$ represent the height and width of the input image, and $\alpha$ denotes the upscaling factor. Here, we use the residual dense block (RDB) \cite{RDN} as the basic block in our SR module since it can fully use features from all preceding layers to generate hierarchical representations, which is demonstrated beneficial to SR reconstruction \cite{iPASSR}.

Specifically, the input image $\mathcal{I}_{LR}$ is first fed to a $3\times3$ convolution to generate initial feature $\mathcal{F}_0 \in \mathbb{R}^{H\times W\times 64}$. Then, $\mathcal{F}_0$ is fed to 8 cascaded RDBs for deep feature extraction. Within each RDB, we use $5$ convolutions with a growth rate of $32$. As shown in Fig.~\ref{fig:ShipSRDet}(b), features from all the layers in an RDB are concatenated and fed to a $1\times1$ convolution for local fusion. Similarly, features from all the RDBs in our SR module are concatenated for global fusion. Afterwards, the fused feature $\mathcal{F}^{LR}_{out}$ is added with the initial feature $\mathcal{F}_0$ and fed to a sub-pixel layer \cite{PixelShuffle} to generate the upsampled feature $\mathcal{F}^{HR}_{out}$. Finally, $\mathcal{F}^{HR}_{out}$ is fed to a $3\times3$ convolution to produce the residual prediction which is further added with the bicubicly upsampled input image to generate the final SR image $\mathcal{I}_{SR}$.

\subsection{Detection module}
Single shot multi-box detector (SSD) \cite{SSD} is used as the detection module in our \textit{ShipSRDet}. In our detection module, VGG-16 network \cite{VGG16} is used to extract multi-scale features from the input SR image $\mathcal{I}_{SR}$. Simultaneously, feature adaption is performed to integrate features from the SR module with those extracted by the VGG network. Specifically, two transition convolutions are performed on $\mathcal{F}^{LR}_{out}$ and $\mathcal{F}^{HR}_{out}$ to adjust their feature depth to 256 and 64 to produce feature $\mathcal{F}^{LR}_{SSD}$ and $\mathcal{F}^{HR}_{SSD}$, respectively. Then, $\mathcal{F}^{HR}_{SSD}$ and $\mathcal{F}^{LR}_{SSD}$ are added to the features in the $2^{nd}$ and $7^{th}$ layer of the VGG network to achieve feature integration. After initial feature extraction with the VGG network, an encoder is further employed for high-level feature extraction. Finally, seven feature maps of different resolutions are generated and fed into the detector$\&$classifier to predict the location and category candidates. The final prediction is produced by performing fast non-maximum suppression (fast NMS) on the candidates.

\section{Experiments}\label{sec:experiment}
In this section, we first introduce the dataset and implementation details. Then, we present ablation studies to investigate our network. Finally, we present the visual results produced by our \textit{ShipSRDet}.

\subsection{Datasets and implementation details}
We used the HRSC dataset \cite{HRSC} for both training and test. We followed the original dataset split and performed $2\times$ and $8\times$ bicubic downsampling to generate training and test image pairs. Consequently, each HR image has a resolution of $512\times512$ and its corresponding medium-low resolution conterpart has a resolution of $128\times128$. Random horizontal flipping, random vertical flipping and random rotation were performed for data augmentation.

Our \textit{ShipSRDet} was implemented in PyTorch on a PC with an RTX 2080Ti GPU, and trained in a two-stage pipeline. In the first stage, we followed \cite{LF-InterNet,LF-DFnet} to train our SR module using the generated image pairs with an $L_1$ loss. Adam method \cite{Adam} is used for optimization. The batch size was set to 4 and the learning rate was was initially set to $1\times10^{-4}$ and halved for every 200 epochs. The training was stopped after 450 epochs. In the second stage, we concatenated the SR module with the pre-trained detection module\footnote{We used the publicly available SSD network which was pre-trained on the COCO \cite{COCO} dataset.}, and performed end-to-end finetuning for global optimization. In the finetuning stage, the learning rate was initially set to $1\times10^{-4}$ and decreased by a factor of 0.1 for every 10 epochs. The finetuning process was performed for 24 epochs.

For evaluation, we followed \cite{voc2012} to use the mean average precision (mAP) as the quantitative metric with the Intersection over Union (IoU) being set to $0.5$.

\subsection{Ablation Study}
We compare our \textit{ShipSRDet} with several variants to investigate the potential benefits introduced by our network modules and design choices. Here, we validate the effectiveness of our method by introducing the following three variants:
\begin{itemize}
    \item \textbf{Bicubic+SSD}: We use the bicubic interpolation approach to upsample the input image, and use the SSD method for ship detection. This variant is used as a baseline method for comparison.
    \item \textbf{SRnet+SSD}: We use the pretrained SR module (i.e., SRnet) to super-resolve the input image, and perform ship detection on the super-resolved images by using the SSD method. In this variant, image SR and ship detection are performed separately without end-to-end finetuning.
    \item \textbf{(SRnet+SSD)\_ft}: We canceled the feature integration in our \textit{ShipSRDet} and perform end-to-end finetuning on the variant \textit{SRnet+SSD}. In this way, only super-resolved images are transferred from the SR module to the detection module for ship detection.
\end{itemize}

\begin{table}
\centering
\renewcommand\arraystretch{1.1}
\small
\caption{Comparisons of the mAP and average running time (Avg time) achieved by different variants of our network. Note that, Avg time is calculated based on an input image of size 128$\times$128. `\_ft' denotes end-to-end finetuning.} \label{tab:ablation}
\vspace{0.2cm}
\begin{tabular}{|l|c|c|}
\hline
Model                   &  mAP  &      Avg time  \\
\hline
Bicubic$+$SSD           & 58.80 \%      & 81 ms  \\
\hline
SRnet$+$SSD             &  60.10 \%     & 163 ms \\
\hline
(SRnet$+$SSD)\_ft       &  63.80 \%     & 163 ms \\
\hline
\textit{ShipSRDet} (proposed)  &   64.50 \%  &  190 ms \\
\hline
HR$+$SSD                &  68.80 \%     &  81 ms \\
\hline
\end{tabular}
\end{table}

 Table~\ref{tab:ablation} shows the comparative results achieved by our \textit{ShipSRDet} and its variants. It can be observed in the table that using the super-resolved images as the input of SSD,  \textit{SRnet+SSD} achieves an improvement of $1.30\%$ in mAP as compared to \textit{Bicubic+SSD}. It demonstrates that the details recovered by the SRnet are contributive to the detection performance to some degree. Note that, a further $3.70\%$ improvement in mAP can be achieved if end-to-end finetuning is performed on the variant \textit{SRnet+SSD}. That is because, by performing end-to-end finetuning, the SRnet can successfully learn to super-resolve an image in a detection-driven manner. The above experimental results also demonstrate that the end-to-end finetuning is significantly beneficial to the overall detection performance. As compared to \textit{(SRnet+SSD)\_ft} which only transfers an image from the SR module to the detection module, our proposed \textit{ShipSRDet} can further achieve an improvement of $0.70\%$ in mAP by integrating feature representations of these two modules, and approximates the upper bound (i.e., $68.80\%$) produced by performing SSD on the HR image. This clearly demonstrates that the super-resolved feature representation is beneficial to the performance of ship detection.

 Besides the detection accuracy, we also report the average running time in Table~\ref{tab:ablation}. As shown in the table, directly performing an SSD detector on the HR image takes $0.81$ ms while using our SR module as a pre-processing step will lead to an increase of 0.82 ms in average running time. When the feature representation is transferred in the two modules, the average running times achieves $190$ ms. That is because, the feature integration operation decreases the parallel processing capability of the network. In summary, our \textit{ShipSRDet} can achieve improved detection accuracy at the cost of minor decrease of the efficiency.

\begin{figure}[t]
\centering
\includegraphics[width=8.3cm]{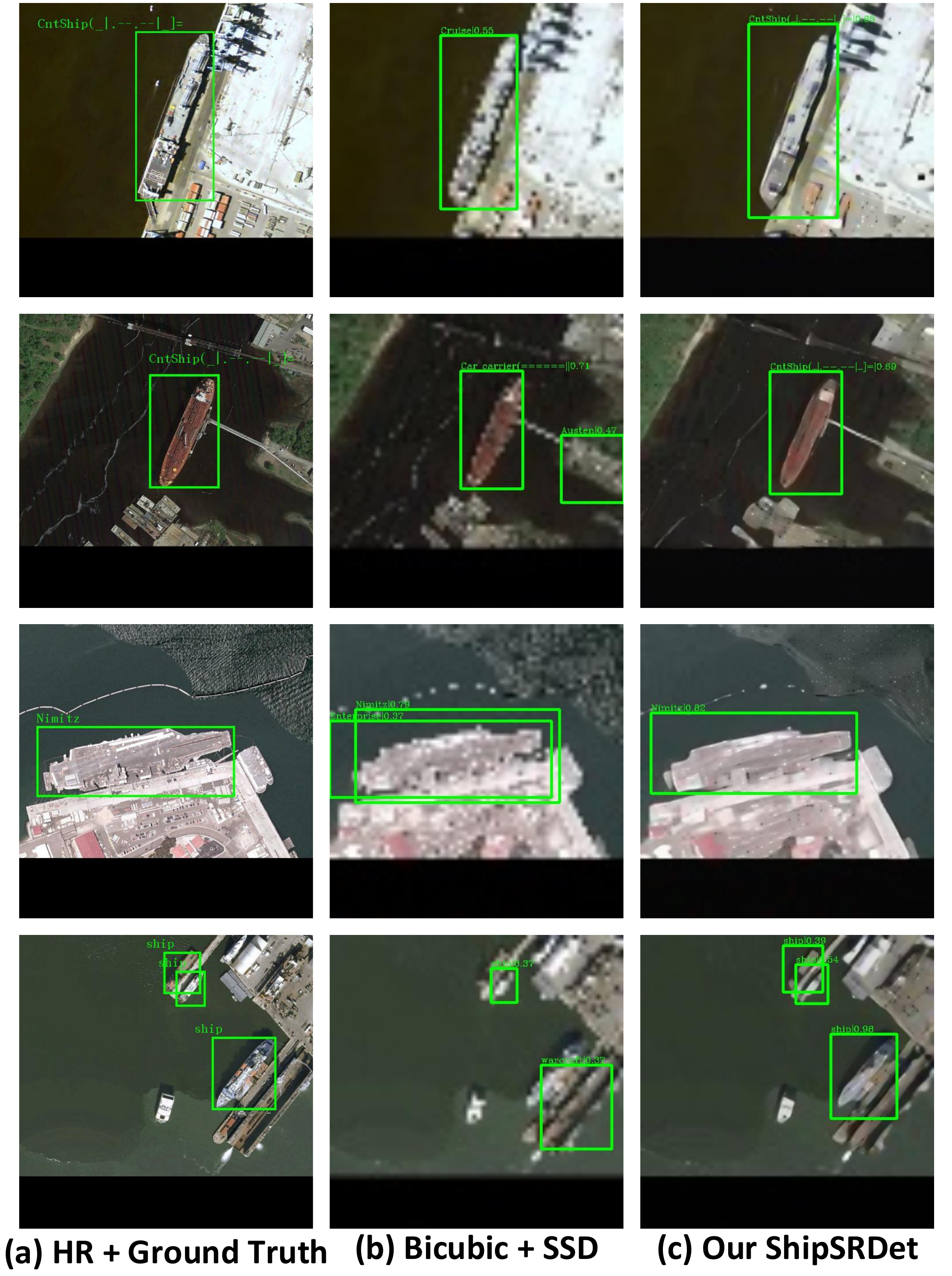}
\caption{Visual results achieved by our \textit{ShipSRDet} on the HRSC dataset \cite{HRSC}.} \label{fig:visual}
\end{figure}

\subsection{Visual Results}
Visual results achieved by the bicubic interpolation approach and our \textit{ShipSRDet} are shown in Fig.~\ref{fig:visual}. As compared to bicubicly interpolated images, the images generated by our network have finer details and are more faithful to their HR groundtruths. Since more details are provided in the super-resolved images and their feature representations, our network achieves a superior detection performance. In contrast, directly performing SSD detectors on the bicubicaly upsampled images will lead to miss detections (the $4^{th}$ row in Fig.~\ref{fig:visual}) and false alarms (the $2^{nd}$ and $3^{rd}$ rows in Fig.~\ref{fig:visual}).

\section{Conclusion and Future Work}\label{sec:conclusion}
In this paper, we propose a \textit{ShipSRDet} to super-resolved remote sensing images for ship detection. Experimental results have demonstrated that both reconstructed high-resolution images and their feature representations are beneficial to ship detection. In the future, we will apply this scheme to more state-of-the-art detectors, and validate its effectiveness on more challenging scenarios such vehicle detection and tiny person detection.

\section*{Acknowledgement}
This work was supported by the National Natural Science Foundation of China under Grant 62071474.

\bibliographystyle{abbrv}
\bibliography{ShipSRDet}

\begin{thebibliography}{10}

\bibitem{courtrai2020small}
L.~Courtrai, M.~Pham, and S.~Lef{\`e}vre.
\newblock Small object detection in remote sensing images based on
  super-resolution with auxiliary generative adversarial networks.
\newblock {\em Remote Sensing}, 12(19):3152, 2020.

\bibitem{dong2020remote}
X.~Dong, L.~Wang, X.~Sun, X.~Jia, L.~Gao, and B.~Zhang.
\newblock Remote sensing image super-resolution using second-order multi-scale
  networks.
\newblock {\em IEEE T-GRS}, 2020.

\bibitem{voc2012}
M.~Everingham and J.~Winn.
\newblock The pascal visual object classes challenge 2012 (voc2012) development
  kit.
\newblock {\em Pattern Analysis, Statistical Modelling and Computational
  Learning, Tech. Rep}, 8, 2011.

\bibitem{Adam}
D.~P. Kingma and J.~Ba.
\newblock Adam: A method for stochastic optimization.
\newblock {\em ICLR}, 2015.

\bibitem{COCO}
T.~Lin, M.~Maire, S.~Belongie, J.~Hays, P.~Perona, D.~Ramanan, P.~Doll{\'a}r,
  and C.~Zitnick.
\newblock Microsoft coco: Common objects in context.
\newblock In {\em ECCV}, pages 740--755, 2014.

\bibitem{SSD}
W.~Liu, D.~Anguelov, D.~Erhan, C.~Szegedy, S.~Reed, C.~Fu, and A.~Berg.
\newblock Ssd: Single shot multibox detector.
\newblock In {\em ECCV}, pages 21--37, 2016.

\bibitem{HRSC}
Z.~Liu, L.~Yuan, L.~Weng, and Y.~Yang.
\newblock A high resolution optical satellite image dataset for ship
  recognition and some new baselines.
\newblock In {\em ICPR}, volume~2, pages 324--331, 2017.

\bibitem{rabbi2020small}
J.~Rabbi, N.~Ray, M.~Schubert, S.~Chowdhury, and D.~Chao.
\newblock Small-object detection in remote sensing images with end-to-end
  edge-enhanced gan and object detector network.
\newblock {\em Remote Sensing}, 12(9):1432, 2020.

\bibitem{shermeyer2019effects}
J.~Shermeyer and A.~Etten.
\newblock The effects of super-resolution on object detection performance in
  satellite imagery.
\newblock In {\em CVPR Workshop}, 2019.

\bibitem{PixelShuffle}
W.~Shi, J.~Caballero, F.~Husz{\'a}r, J.~Totz, P.~Aitken, R.~Bishop,
  D.~Rueckert, and Z.~Wang.
\newblock Real-time single image and video super-resolution using an efficient
  sub-pixel convolutional neural network.
\newblock In {\em CVPR}, pages 1874--1883, 2016.

\bibitem{VGG16}
K.~Simonyan and A.~Zisserman.
\newblock Very deep convolutional networks for large-scale image recognition.
\newblock {\em arXiv:1409.1556}, 2014.

\bibitem{LF-InterNet}
Y.~Wang, L.~Wang, J.~Yang, W.~An, J.~Yu, and Y.~Guo.
\newblock Spatial-angular interaction for light field image super-resolution.
\newblock In {\em ECCV}, pages 290--308, 2020.

\bibitem{LF-DFnet}
Y.~Wang, J.~Yang, L.~Wang, X.~Ying, T.~Wu, W.~An, and Y.~Guo.
\newblock Light field image super-resolution using deformable convolution.
\newblock {\em IEEE T-IP}, 30:1057--1071, 2021.

\bibitem{iPASSR}
Y.~Wang, X.~Ying, L.~Wang, J.~Yang, W.~An, and Y.~Guo.
\newblock Symmetric parallax attention for stereo image super-resolution.
\newblock {\em arXiv:2011.03802}, 2020.

\bibitem{RDN}
Y.~Zhang, Y.~Tian, Y.~Kong, B.~Zhong, and Y.~Fu.
\newblock Residual dense network for image super-resolution.
\newblock In {\em CVPR}, pages 2472--2481, 2018.

\end{thebibliography}

\end{document}